# Insect-Computer Hybrid System for Autonomous Search and Rescue Mission

P. Thanh Tran-Ngoc[1], D. Long Le[1], Bing Sheng Chong[1], H. Duoc Nguyen[1], V. Than Dung[1], Feng Cao[1], Yao Li[2], Kazuki Kai[1], Jia Hui Gan[1], T. Thang Vo-Doan[3], T. Luan Nguyen[4], and Hirotaka Sato[1*]

[1]School of Mechanical & Aerospace Engineering, Nanyang Technological University; 50 Nanyang Avenue, 639798, Singapore

[2]School of Mechanical Engineering and Automation, Harbin Institute of Technology, Shenzhen; University Town, Shenzhen, 518055, China

[3]University of Freiburg; Hauptstrasse. 1, Freiburg, 79104, Germany

[4]University of Leeds; Woodhouse, Leeds LS2 9JT, United Kingdom

*Corresponding author. Email: hirosato@ntu.edu.sg

**Abstract:** There is still a long way to go before artificial mini robots are really used for search and rescue missions in disaster-hit areas due to hindrance in power consumption, computation load of the locomotion, and obstacle-avoidance system. Insect–computer hybrid system, which is the fusion of living insect platform and microcontroller, emerges as an alternative solution. This study demonstrates the first-ever insect–computer hybrid system conceived for search and rescue missions, which is capable of autonomous navigation and human presence detection in an unstructured environment. Customized navigation control algorithm utilizing the insect's intrinsic navigation capability achieved exploration and negotiation of complex terrains. On-board high-accuracy human presence detection using infrared camera was achieved with a custom machine learning model. Low power consumption suggests system suitability for hour-long operations and its potential for realization in real-life missions.

**INTRODUCTION**

In an Urban Search and Rescue (USAR) mission, fast pinpointing of victim locations trapped by structural collapse is crucial to maximize the chance of survival. The time-consuming process of victim extraction would only be done after confirming the possible victim locations through reconnaissance, which takes 3 h on average [1]. However, as the locations of the trapped victims are usually inaccessible to humans, various tools such as electronic cameras, seismic sensors, and sniffing dogs are utilized to track down possible hidden victims in a small area. Yet the current methods brought along constraints such as handling difficulty, limited range, sensitivity to environment, and slow searching speed, which highly limit the number of victims that can be quickly located [2]. With respect to this, mini robots with dimensions of 10 cm or less can be immensely helpful in increasing search efficiency. There are certain requirements for such robots: they should be lightweight and small enough to penetrate through the rubble and locate the victims inside. High level of mobility is necessary to autonomously negotiate and traverse through obstacles even without commands from a remote operator. The ability to detect a human life is also required to locate survivors automatically. Finally, wireless communication is also necessary to report to a remote operator or a rescue team base. [2]

Owing to breakthroughs in fabrication technologies and actuator developments, researchers have built mini robots that can walk/crawl [3-5], climb [6], or fly [7], which are compatible with USAR missions. However, there are no mini robots capable of autonomous navigation inside an unknown environment, mainly because of large physical dimensions and high power requirements [8]. For example, the simultaneous localization and mapping (SLAM) technique [9], commonly used in robot navigation, relies on bulky and/or resource-consuming devices such as RGB-D (Red Green Blue–Depth) camera [10], LIDAR (Laser Imaging Detection and Ranging) [11], and CMOS (Complementary

Metal-Oxide Semiconductor) camera [12,13]. It follows that the proposed obstacle-negotiation solutions [14,15] usually require large and complex mechanical systems to avoid obstacles while traveling across uneven terrain. Furthermore, as the on-device battery capacity is usually limited to around 100 mAh, the large locomotion power consumption in the order of 10–1,000 mW only hampers the performance of mini robots by depleting the power source within minutes [3-5]. The total power consumption must be controlled by compromising weight, size, and power usage in selecting robot components and sensors.

Human detection is another important criterion for USAR robots. Due to the strict size, power, and resource limitations, a human detection system deployed in mini robots may not be able to use all state-of-the-art human detection methods employing visible-light cameras, infrared (IR) cameras, or UWB (Ultra-Wide Band) radars. Multiple previous works have been done in this area to develop detection algorithms usable for low-power systems [16,17]. Based on these works, mini robots with the ability to detect human or object were further explored, but their limitations such as low accuracy of sensors [18], lack of on-board computational resources to run the algorithm [19], and large power consumed by the locomotion system [20] have prevented their practical applications. Therefore, it is evident that creation of mini robots satisfying all the USAR requirements is almost impossible. It is immensely difficult to implement high-performance components, such as processors, sensors, and wireless module on the robot, considering that the locomotion actuators already consume most of the energy from the on-device power source.

An alternative solution for the abovementioned issues has emerged over the past decade, in the form of insect–computer hybrid systems, so-called cyborg insects, or biobots. These hybrid systems are the fusion of a living insect platform and a miniature electronic controller, thus showing both the locomotion proficiency of an insect and the controllability of a robot. With their low mass and volume, further enhanced by their intrinsic ability to walk, fly, and sense the surrounding environment, insect–computer hybrid systems can easily achieve the task of penetrating and navigating inside complex environments to search for a target [21,22]. Different hybrid systems with different movement capabilities have been introduced in the past, for example, moth or giant flower beetle with flight initialization, cessation, steering, and thrust regulation [23-27], or cockroach and beetle with walking control for ground operation [28-31]. Owing to their natural ability to maneuver through obstacles, the insect–computer hybrid system requires much simpler hardware and control algorithm to operate in a complicated unknown terrain compared with their artificial counterparts [7,32]. In addition, as the insect platform is mainly controlled via electrical stimulus at neural, neuromuscular, or sensory sites [21,27,28,30], their locomotion control can be achieved with just a few 100 µW power [29,31]. This leads to comparatively lower power consumption in insect–computer hybrid systems and allowing their participation in operations that last for an hour or more.

In this study, we propose a comprehensive solution for USAR using an insect–computer hybrid system with obstacle negotiation and human detection capability (Fig. 1a). A Madagascar cockroach was used for the hybrid system, which has a maximum payload limit of 15 g [21] and an average size of 6 × 2 cm. A single backpack circuit board was mounted on the insect, which included insect locomotion control system, positioning and navigation system, miniature IR camera, and wireless communication module. A stimulation protocol and a simple feedback navigation program was developed and implemented to utilize the insect's intrinsic obstacle-negotiation ability in motion control (Fig. 1b). Early testing of the system showed promising success in navigating through unknown environments toward a destination, although a more complex environment can cause potential navigation failures to occur. To overcome these failures, a new navigation program, named as Predictive Feedback navigation, was developed to prevent the occurrence of failure scenarios by promptly directing the insect out of the risky locations (Fig. 1b). Besides that, an image classification model that was based on Histogram of Oriented Gradients and Support Vector Machine was loaded on the board to achieve autonomous human detection. The whole system was subsequently tested in a mock-disaster scenario to evaluate its performance (Supplementary Video 1). The completed hybrid system has a compact size and can navigate to predetermined destinations, autonomously traverse unknown obstacle-terrain, detect human presence, and wirelessly report it to a remote console. We will adopt this technology for assisting USAR missions in disaster-hit areas in the near future.

**RESULTS AND DISCUSSION**

*Navigation of Insect–computer Hybrid System with Simple Feedback Control Algorithm*

The insect–computer hybrid system consists of an insect platform manipulated by a customized wireless backpack control system (Fig. 1a). The insect's movement was directed by electrically stimulating its left and right cerci to induce turning (Extended Data Fig. 1a); when the left cercus was stimulated, the insect made a right turn (clockwise rotation), and vice versa (Supplementary Video 2). Based on this stimulation protocol, a Simple Feedback control algorithm (Fig. 2) was developed to enable the automatic navigation of the insect in a predefined area (Extended Data Fig. 2). This algorithm operated with no prior knowledge of the environment and steered the insect according to the error between the insect orientation and the direction to user-defined destination (Extended Data Fig. 3a, Extended Data Table 1). It was deployed in the on-board system, which was then utilized to demonstrate the successful insect navigation to a predetermined destination in an environment without obstacles (Fig. 3a, Extended Data Fig. 2a, 4a).

Under the presence of low obstacles (with height of 1.5 cm, Extended Data Fig. 2b), it was still able to reach the destination (Extended Data Fig. 3b) in 48 out of 49 trials (Fig. 3A), showing that it could retain the intrinsic obstacle-negotiation ability even while being stimulated. In 35 success trials, the insect simply climbed over the first obstacle that it encountered (Extended Data Fig. 3d, 4b) in an orthogonal manner, i.e., the angle $\theta$ between the insect's orientation and the nearest obstacle (Extended Data Fig. 3c) was 74.89 ± 10.24 degrees. In the other 12 successful trials, the insect moved along the edge of the obstacle before climbing over from the side (Extended Data Fig. 4b), i.e., $\theta$ was 38.77 ± 6.54 degrees. There was only one instance where the insect was navigated along the edge of the obstacle without any climbing motion (Extended Data Fig. 4b). Evidently, the hybrid system demonstrated superior capability of handing obstacles, with height comparable to its own, without any prior information; similar ability from artificial mini robots is still far from reality. Regardless of the obstacle-negotiation strategy used, once the insect overcame the obstacle, the navigation program could successfully navigate it to reach the destination.

A different obstacle-negotiation behavior was observed when taller obstacles (walls with height of 10 cm, Extended Data Fig. 2c) were placed in the arena. When the insect was being steered toward a wall, it could not climb over and had the tendency to align its body along the wall, i.e., the angle $\theta$ became small (Extended Data Fig. 5a–b). The insect then readjusted itself by performing a backward motion while staying pressed against the wall due to stimulation (Fig. 3d, Extended Data Fig. 5b-c), allowing it to move away from the obstacle such that it might continue its navigation without being blocked. However, with the presence of tall obstacles, the Simple Feedback navigation frequently failed to navigate the insect to the destination before the experiment was terminated, specifically in 37 out of 49 trials (Fig. 3a, 3e). Out of all trials, the insect's immobility was observed 17 times and 16 times, with and without the presence of stimulation, respectively (Fig. 3c). There were four trials that were categorized as timeout failure, in which the insect moved continuously but was still unable to reach the destination within a set time (Fig. 3d).

*Limitation of the Simple Feedback Navigation in the Tall Obstacles Environment*

There were 17 out of 37 failed navigation trials in which the insect was immobilized while under stimulation. At the locations where the insect became immobile, the $\theta$ was small, suggesting that it was physically blocked by the wall (Fig. 3c, Extended Data Fig. 5a). However, as the insect was not facing directly toward the destination, the navigation program maintained the stimulation even though it could not move further.

Contrarily, 16 of 37 failed trials were accounted to the insect's immobilization without electrical stimulation. In these cases, the insect's obstacle-negotiation behavior when encountering a wall cooperated with the steering stimulation from the navigation program, allowing the insect to successfully reorient itself to face toward the target, causing the program to deactivate the stimulation (Fig. 3c, Extended Data Fig. 5b). However, the insect did not voluntarily continue to move after the stimulation was deactivated, thereby failing to reach the destination before the experiment was terminated.

A minority of 4 out of 37 failed trials belonged to the situation wherein the obstacle negotiation ability of the insect negatively affected its navigation process. After the insect retreated when encountering a wall, the navigation program stimulated it to move toward the obstacle again, and the whole process was repeated (Fig. 3d). As a result, the insect was trapped in that local vicinity until timeout occurred. This contradiction between the insect's obstacle-negotiation behavior and the navigation program was the direct cause of all four timeout trials, where nearly half of the total navigation time was spent on backward motion, i.e., 40.93 ± 13.34 s out of 100 s (Extended Data Fig. 5c).

*Enhanced Navigation Performance with the Predictive Feedback Control Algorithm*

The main reason that the Simple Feedback navigation algorithm failed was its inability to recognize and handle the immobility or "trapped state" (in the case of timeout failures) of the insect. To overcome these shortcomings, a new double feedback control navigation system, henceforth named as "Predictive Feedback" navigation (Fig. 2), was developed to improve the navigation performance in the tall obstacles situation while retaining the original performance in environments with zero or low obstacles. Since the insect typically walked straight without any external stimulus [33,34], the new program could predict its immobility when not under steering stimulation by monitoring its linear speed ($v_l$, Extended Data Fig. 3e), then preventing the motion stop by prematurely accelerating the insect when $v_l$ decreased below a given threshold (Fig. 1b, Extended Data Table 1). Similarly, both cases where the insect was immobile under steering stimulation and the timeout cases could be prevented by monitoring the insect's angular speed ($\omega$, Extended Data Fig. 3e) while steering it. A small measured $\omega$ (Extended Data Table 1) would indicate that the insect might be blocked by an obstacle, and it would otherwise cause the insect to either stay immobile or start a backward motion if not intervened. The new navigation program hindered this behavior by interrupting the original stimulation and accelerating the insect away from the local vicinity (Fig. 1b). The acceleration was attained by stimulating the insect's two cerci simultaneously (Supplementary Video 2) [35].

This Predictive Feedback navigation program achieved 100% success rate in both environments with no obstacle and with only low obstacles (Fig. 3a, Extended Data Fig. 4c-d), meaning it retained the original performance of the Simple Feedback navigation in such conditions. This can be clearly seen as the insect in both navigation systems displayed the major trend of climbing over low obstacles, i.e., this behavior had 96% rate of occurrence (Extended Data Fig. 4b, 4d). The angle $\theta$ when climbing occurred in the new navigation system was 75.10 ± 10.68 degrees, which was also comparable to the previous navigation system. These similarities in responses resulted in a minor difference in navigation time, specifically 8.29 ± 6.29 s in Simple Feedback navigation and 6.35 ± 4.92 s in Predictive Feedback navigation (Fig. 3b).

A higher success rate of 94% was recorded with the new navigation program in situations with tall obstacles (Fig. 3a, 3f). As shown in a representative case (Fig. 4), $\omega$ and $v_l$ of the insect were periodically measured during stimulation and free walking, respectively. In this example, there were four locations where $\omega$ and $v_l$ were below their thresholds, each leading to an acceleration event. In the first and fourth events, the insect was walking freely while facing the destination when the $v_l$ dropped below its given threshold. As a result, the system accelerated the insect to keep it in motion. The second and third events occurred when $\omega$ was measured to be smaller than its threshold due to the insect being steered toward an obstacle. The stimulation was then interrupted by the navigation program, and the insect was accelerated so that it rapidly escaped the vicinity, preventing it from stopping or being trapped in the area. It is observed that during the acceleration in the first and fourth events, the insect dashed toward the obstacle and moved along the wall. This behavior was consistent with present studies on the insect's response to the obstacles situated on its escape path [36,37]. Such a reaction allows the navigation program to take advantage of the insect's natural obstacle negotiation reaction, which would be beneficial for cases where obstacle detection/crash prevention sensors could not be used.

The ability of the Predictive Feedback navigation program to predict and prevent immobility of the insect could be observed by comparing the recorded angles $\theta$ in the two navigation systems, in which there was a high correlation between the locations of acceleration events in the Predictive Feedback navigation program and the locations of the insect's immobility in the Simple Feedback navigation program (Fig. 3c, Extended Data Fig. 6). Also, the insect controlled by the Predictive Feedback navigation program spent $5.15 \pm 5.33$ s in backward motion, which was eight times lesser than the previous navigation program. With the reduction in time spent for backward motion, the new navigation program improved the success rate by reducing probability of timeout, and generally shortened the total navigation time from $48.98 \pm 19.49$ s to $33.91 \pm 18.35$ s (Fig. 3b). This shows that the new navigation program successfully overcame the tall-obstacle negotiation issue. When implemented into the insect-computer hybrid system, a mini-scale solution for the navigation challenges in obscured environment can be achieved.

Nevertheless, it should be noted that there were three failed trials, i.e., 6% out of all trials, recorded for the Predictive Feedback navigation, which was caused by the insect missing the obstacle entrance due to acceleration and thereby forcing the insect to navigate around the obstacles again until timeout occurred (Fig. 3f). However, this overshooting also occurred in other successful trials (Fig. 3f), implying that these failed trials can be easily solved by increasing the experiment duration. With this change, the insect will have sufficient time to reach the destination.

*On-board Human Detection System*

To achieve the goal of autonomous search and rescue in structural collapse disasters, the insect–computer hybrid system must be equipped with an accurate human detection system. This detection system must be able to operate under a lightless condition for at least several hours during a searching mission. A continuous data streamline from the system to its control station, thus, would not be applicable due to high energy consumed by the wireless communication. All decisions in detecting human presence, therefore, should be performed independently by the system, which was constrained due to hardware size, power, and computational resource. In this study, we successfully designed and developed a fast, precise, and energy-efficient human detection system, consisting of a low-powered IR camera coupled with an on-board machine learning algorithm to recognize human signature.

Owing to the difference between human body temperature and the ambient temperature, IR camera was often used to detect humans [38]. This type of camera would be suitable for search and rescue activities under rubbles, where a normal camera would not function well without a proper light source. The developed human detection system utilized an IR thermopile array with $32 \times 32$ pixels and $90° \times 90°$ field of view (Extended Data Fig. 1c) to collect IR images of the surroundings. The main microcontroller denoised these images by passing them through a median filter [39] before outputting the human presence results, which were calculated via a custom human detection algorithm (Fig. 5A). This algorithm employed the commonly-used Histogram of Oriented Gradients (HOG) and Support Vector Machine (SVM) [17] (Fig. 5a), which was built and tested on a computer to determine the suitable parameters before being deployed as the on-board function. A HOG feature descriptor with a cell size of $4 \times 4$ was selected after comparing the algorithm's accuracy with different cell sizes (Extended Data Table 2), and a linear SVM was used for the classifier due to its fast yet still sufficiently accurate classification results (Fig. 5b–c).

The established human detection algorithm achieved 87% average accuracy in classifying between human and non-human subjects (Fig. 5b), comparable to other studies on human classification using HOG and linear SVM [17,40]. Specifically, 90% average recall rate can be achieved for human images captured within the distance between 0.5 m and 1.5 m (Extended Data Fig. 7). In addition to the high accuracy, the developed model only occupied 18.3 kB (~1%) of Flash and 52.2 kB (~20%) of Static Random Access Memory (SRAM) during on-board implementation (Extended Data Table 3), and its average computational time was only 95 ms. This low processing time allowed the system to collect and classify many IR images during the insect's navigation, which intuitively would reduce the probability of overlooking the entrapped victims. Although the detection frame rate is lower compared with similar systems used to detect pedestrians [41], it was sufficient to preserve the locality of each classified image under the average insect speed of 3 cm/s [28].

Without platform limitations, similar human detection systems were implemented in Field-programmable gate arrays (FPGA) [41], which were designed to have higher computational capability and data process rate with higher power cost. For example, 566 mW of power was consumed just to run the linear classification on a Virtex 7 platform [17]. In comparison, the energy consumed by our human detection algorithm running at 100% duty cycle was only 24 mW when embedded on the current microcontroller. With a total system power consumption of 205.5 mW, the insect–computer hybrid system could operate for up to 2.2 hours using a 120 mAh battery, suitable for a practical USAR mission.

*Insect–computer Hybrid System Searching for Human Autonomously in an Unknown Environment*

A demonstration of the insect–computer hybrid system in search and rescue operation was performed using the Predictive Feedback navigation algorithm along with the on-board human detection system. An insect–computer hybrid system was navigated autonomously through a mock-disaster site, with the inclusion of several human and non-human objects to evaluate its detection performance (Fig. 5d, Supplementary Video 1).

Despite the lack of obstacles information, the Predictive Feedback navigation algorithm successfully guided the insect to reach all the given targets in a predetermined sequence. During the navigation demonstration, the insect performed its obstacle negotiation ability by either climbing over the cement fragments or following the contour of obstacles, both with and without electrical stimulations (Fig. 5d). There were seven times the insect was accelerated when the inertial measurement unit (IMU) measured its angular speed to fall below the threshold, and two acceleration events were generated when the IMU measured a low linear speed.

During the whole demonstration, the insect–computer hybrid system actively searched for humans. In the demonstration, three hot objects varying in shape and sizes, and two humans were placed on the insect's travel path. The on-board human detection system was able to distinguish between these subjects, which can be seen as it correctly classified the human for 11 times while ignoring the non-human hot objects (Fig. 5e). This result demonstrated the feasibility of the insect–computer hybrid system in search and rescue missions where it must navigate into an unknown environment while scanning the area to locate victims.

**CONCLUSION**

We have presented the proof of concept and demonstration of the first-ever mini-scale insect–computer hybrid solution for USAR, which overcomes many flaws of USAR mini robots in terms of maneuverability and power consumption. The hybrid system is capable of navigating and searching for victims in unknown environments, and has a remarkably compact size yet exhibited a robust obstacle handling capability due to the custom navigation program that retains the natural locomotive ability of the insect. The inclusion of an onboard human detection system using IR images removes the need for manual control, allowing the hybrid system to work autonomously and improve the overall searching efficiency. In addition, the low power consumption of the whole system would allow for few hours of operation time. This study successfully proved the efficiency of a biological machine implementation in solving the issue of power-hungry actuators in artificial robots, thereby allowing the saved energy to be better utilized for other tasks, such as sensing and communication. Further improvements can be implemented to improve its searching efficiency, such as an accurate on-board localization system allowing real-time position tracking of the insect would enhance the response speed of the rescue team once the insect–computer hybrid system finds a victim inside the rubble. Solutions such as dead reckoning using IMU or UWB localization can be implemented without heavy load on power supply or computational resources [42]. Besides, the capabilities of the system could also be enhanced with a more efficient power source, such as deploying an integrated biofuel power source [43].

**Acknowledgments**
The authors thank Mr. Seah Hee Chuan and Dr. Goh Bing Hui, Terence at KLASS Engineering & Solutions Pte. Ltd, Mr Cheng Wee Kiang, Mr Ong Ka Hing, and Ms Vanessia Choo at Home Team Science & Technology Agency (HTX), and distinguished officers at Singapore Civil Defence Force (SCDF) for their helpful comments and advices, Mr. Chew Hock See, Mr. Seet Thian Beng, Ms. Kerh Geok Hong Wendy, Mr. Roger Tan Kay Chia, Mr. Li Rui and Mr. Qifeng for their support. This work was supported by KLASS Engineering & Solutions Pte. Ltd (NTU REF 2019-1585).

**Author contributions**
 H. S., P.T.T.N., H.D.N., F.C., T.T.V.D. and T.L.N. conceived and designed the research. H.D.N., P.T.T.N., F.C., K.K. and J.H.G. established insect control protocol. T.L.N., V.T.D., D.L.L., H.D.N., B.S.C., P.T.T.N. and Y.L. developed hardware and software for the backpack. P.T.T.N., B.S.C. and H.D.N. developed the human detection algorithm. P.T.T.N., H.D.N., D.L.L., and B.S.C. conducted the experiment and analysis. P.T.T.N., H.D.N., D.L.L., B.S.C. and H.S. wrote and edited the manuscript. H.S. supervised the research. All authors read and edited the paper.

**Competing interests**
The authors declare no competing interests.

**Additional information**
**Extended data** is available for this paper at XXXXX
**Supplementary information** is available XXXXX
**Correspondence and requests for materials** should be addressed to H.S.

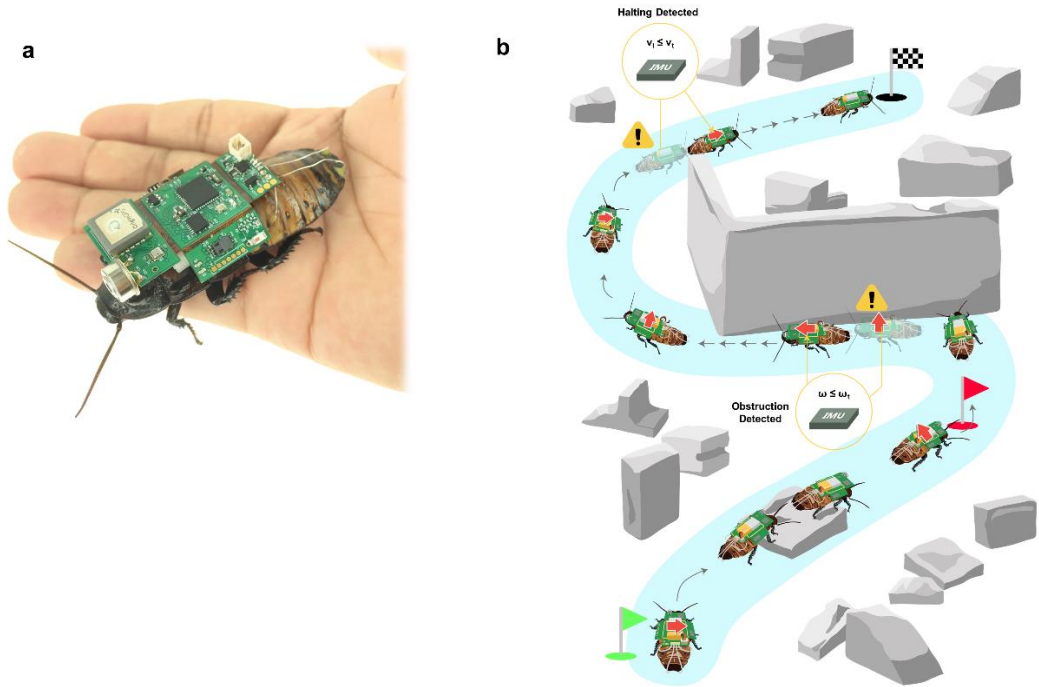

**Fig. 1 | The insect-computer hybrid system. a,** The hybrid system consists of a living Madagascar hissing cockroach and a wireless backpack controller. Autonomous navigation is enabled by electrical stimulating the insect's sensory system, while the infrared camera allows for on-device human detection via image classification. **b,** The established navigation program cooperates harmoniously with the insect's natural locomotive ability to deal with unknown environment during the insect's exploration missions. By controlling the stimulation (denoted with red arrows), the algorithm successfully directed the insect through obstacles to reach predetermined targets. Only data from the on-board inertial measurement unit is used in navigation to predict and prevent potential navigation failure conditions (denoted with warning symbol).

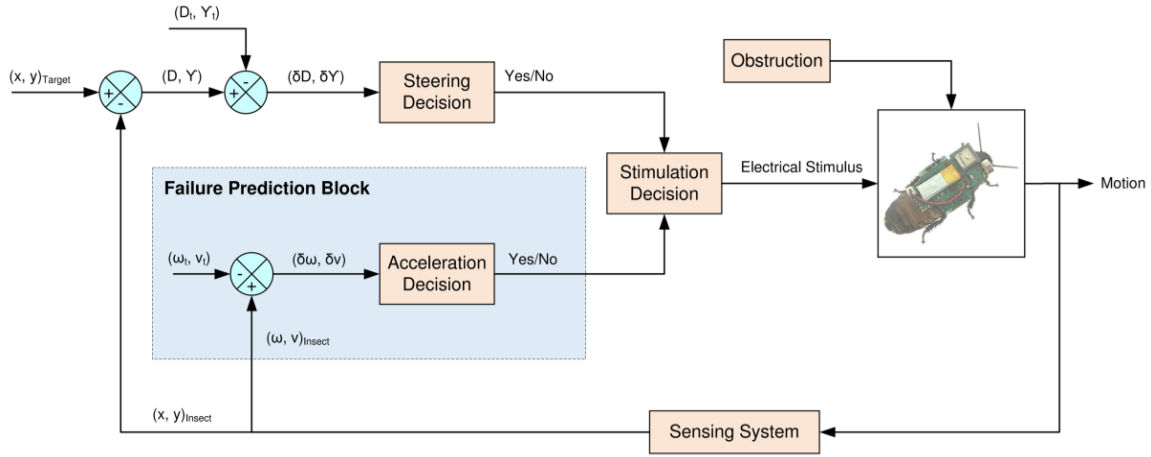

**Fig. 2 | Control diagram of the two navigation programs: Simple and Predictive Feedback Algorithms**. The two navigation algorithms are differentiated by the inclusion of a "Failure Prediction Block," which is exclusively used for the Predictive Feedback algorithm. The Simple Feedback algorithm compares the spatial distance ($D$) and orientation ($\Upsilon$) errors between the target and the insect to their thresholds to initiate either experiment termination ($D \leq D_t$) or insect steering ($\Upsilon \leq \Upsilon_t$). The Predictive Feedback algorithm observes the insect's motion ($\omega$, $v$)$_{Insect}$ to predict failures. If the monitored speeds fell below their thresholds ($\omega_t$, $v_t$) indicating insects is halted by environment interaction, the steering stimulus is immediately discontinued and an acceleration stimulus is released to direct the insect out of its current vicinity, thereby preventing a navigation failure.

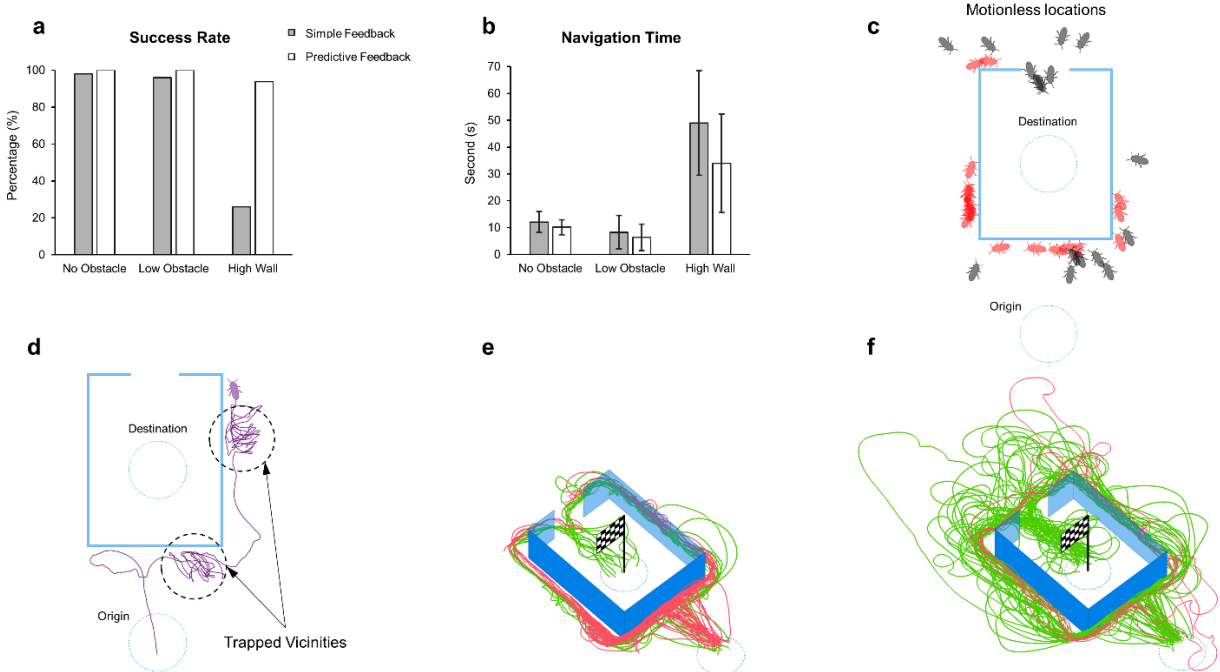

**Fig. 3 | Comparison between the two navigation programs. a,** Success Rate and **b,** Navigation Time for different environments (N = 5 insects, 44 ≤ n ≤ 50 trials for each columns). The two algorithms possess similarly good performances except for the case of tall obstacles, with the Predictive Feedback navigation being superior to its counterpart in terms of success rate and navigation time. **c-d,** Failures in Simple Feedback navigation (N = 5 insects, n = 37 trials). The red, black, or purple colors of insect-like objects each denotes the failures caused by the insect immobility with the presence (17 trials) or absence (16 trials) of electrical stimulus, or timeout failures while remaining in motion (4 trials, only 1 representative trial is shown). The red insects steering motion under stimulation was obstructed by the wall. The black insects successfully oriented their body toward the destination, and the stimulus was terminated, but they simply stopped moving further. The purple insect is trapped in some local vicinity where it performed backward motion to escape an obstruction but then directed back to the obstacle and unable to reach the destination prior to timeout. **e-f,** Trajectories of the insect under the control of simple and Predictive Feedback navigation algorithms, respectively. The Predictive Feedback navigation trials in total displayed significantly less failures (red curves) and more successes (green curves). In addition, insect paths in the successful Predictive Feedback navigations are further away from the walls, implying an effect of the acceleration activated by the Failure Prediction Block.

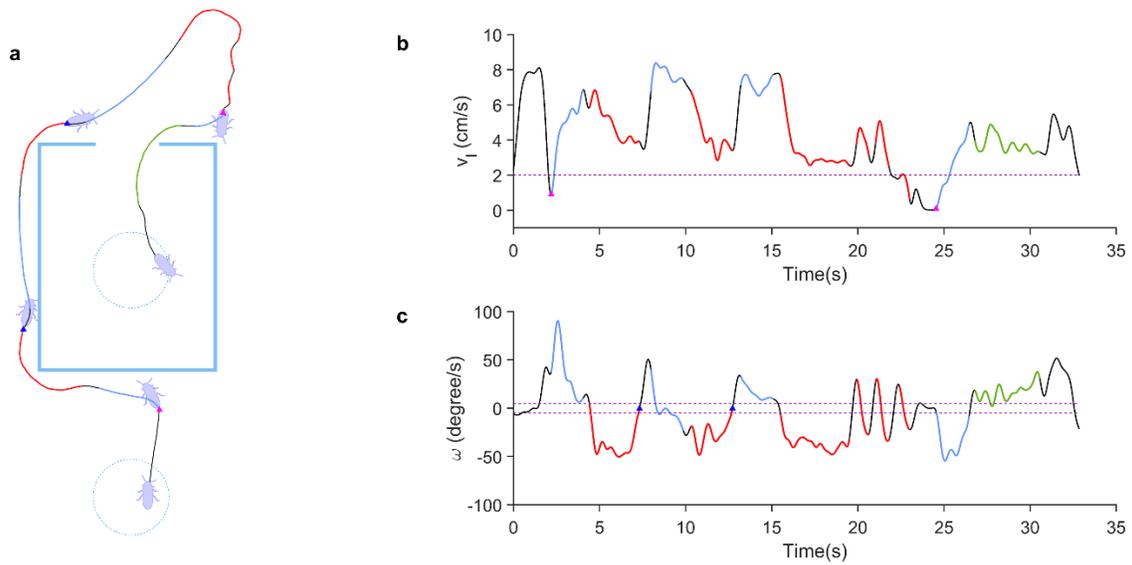

**Fig. 4 | Anatomy of a successful trial attained with the Predictive Feedback navigation. a,** The trajectory, **b-c,** linear ($v_l$) and angular ($ω$) velocities of the insect, respectively. The black, green/red, and cyan segments each represents the free-walking period, left/right steering stimulation, and acceleration, respectively. The purple insect-like objects denote the insect's orientation, and the pink and blue triangles denote positions where $v_l$ and $ω$ fall below their thresholds (i.e., purple dashed lines), respectively. The accelerations fired by the Failure Prediction Block when speeds fell below threshold succeeded in directing the insect out of failure-prone regions and kept it in motion.

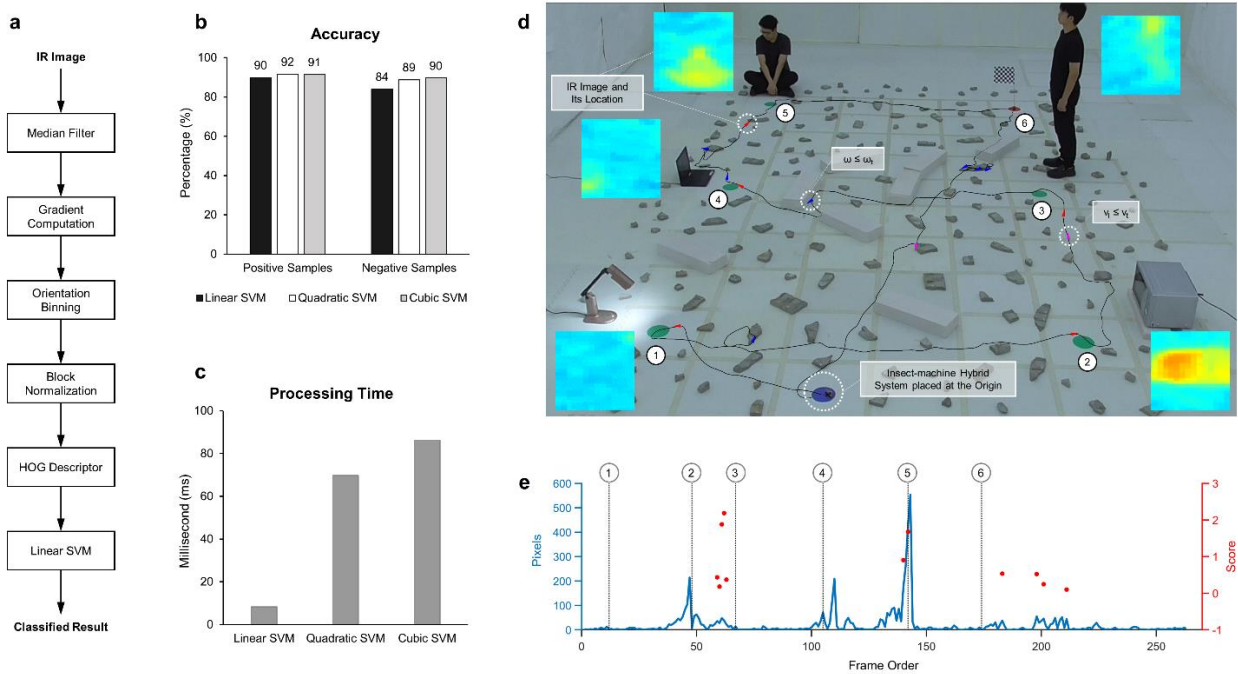

**Fig. 5 | Overview of the developed human detection algorithm and the mock-disaster exploration demonstration. a,** The process of IR image classification to extract feature descriptors and determine human presence in image. **b,** The accuracy and **c,** processing time are compared between three SVM methods. While their detection accuracy is similar, linear SVM pulls ahead in terms of computational simplicity. **d,** In a mock-disaster terrain, the insect was able to reach all destinations (i.e., red/green 8 cm radius circles) in a predefined sequence (shown with the white numbered circles) before returning to the origin. The Failure Prediction Block activations, along with the insect's orientations, are denoted with pink and blue arrows for instances where $v_l$ or $\omega$ fall below thresholds, respectively. The on-board human detection system operated at 1 Hz. Some captured IR images are shown with the insect's corresponding positions indicated via red arrows. **e,** The plot depicts the operation of the human detection algorithm during the demonstration. The algorithm will activate if the pixel count that lies within human temperature range (indicated by the blue curve) were larger than a given threshold (i.e., 15 pixels). The output scores of possible human presences detected by the algorithm are denoted with red dots. The plot is temporally synced with the path in (d), with numbered circles referring to the same numbered locations in (d). The effectiveness of the on-board human detection algorithm can be seen by referring to the corresponding subjects (i.e., either human or hot object), their location and their respective scores along the horizontal axis.

**METHODS**

Animals

Male Madagascar hissing cockroaches (*Gromphadorhina portentosa*, 5.7 ± 0.6 cm, 6.32 ± 1.5 g) were kept in a laboratory terrarium (NexGen® Mouse 500, Allen Town®) and were fed sliced carrots every week. The temperature and relative humidity were maintained at 25 °C and 60%, respectively. The use of the cockroach was permitted by the National Environmental Agency (Permit number NEA/PH/CLB/19-00012).

Electrode implantation and stimulation

Our method of steering the insect's locomotion was based on existing procedures [35,44]. The insect was first anaesthetized using carbon dioxide in an airtight container for 30 s. Sandpaper was used to file the mounting site of the wireless stimulator and the implantation sites on the third abdominal segment, i.e., the black dots located at the sides (Extended Data Fig. 1a). This process allowed the backpack and beeswax to attach to the cuticle securely after implantation. Next, the cerci of the insect and their bases were covered with glue (Extended Data Fig. 1a). After the glue solidified, the tips of the cerci were cut to create a small opening. A Teflon–insulated platinum wire (A–M Systems, coated diameter 0.14 mm, uncoated diameter 0.08 mm) was then inserted coaxially into the cerci, with an implantation depth of 10 mm from the tip of the electrode. The wire was de–insulated using tweezers prior to implantation such that the entire implanted portion was de–insulated. Beeswax was applied to secure the electrode in place and to seal the opening on the cercus.

For the third abdominal segment, an insect pin was used to create a small hole on each implantation site. A platinum wire was then inserted into the third segment while ensuring that the axis of the inserted wire was perpendicular to the surface of the cuticle. The implantation depth was 5 mm from the tip of the electrode. After implantation, beeswax was applied to cover the openings. The other ends of the implanted electrodes were connected to the wireless stimulator, which in turn would be mounted on the dorsal side of the insect throughout the experiments.

The insect was steered by passing current through a pair of the electrode on the same, which induced the insect's rotation in the opposite direction (Supplementary Video 2). The stimulation waveform was a biphasic square waveform of frequency 40 Hz and 50% duty cycle, in the range of 6 – 8 V.

Miniature Wireless Communication Stimulator (Backpack)

The insect-computer hybrid system was controlled by customized wireless circuit board, or a backpack (Extended Data Fig. 1b-c). The system main controller unit was Texas Instruments (TI) MSP432P4011 microcontroller (ARM 32-bit Cortex M4F, 48 MHz, 2MB of Flash, 256 kB of SRAM) which was known for its excellent power efficiency. The stimulation signal was generated by the chip AD5504 (Analog Devices, Quad-channel, 12-bit, 7.3 mV resolution) with voltage supplied up to 12 V using TI TPS61046 Boost Converters. The backpack was integrated with an IMU using MPU9250 (InvenSense, 6 axes used with Digital Motion Processor, 11.55 mW) to detect the hybrid system's movement. For human detection purposes, a HTPA32x32 IR camera was implemented for its small footprint and low power consumption (Heimann Sensor GmbH, 8×8×5 mm, 0.99 g, 21.45 mW average, 8 µW sleep, 32 × 32 resolution, 90° × 90° field of view). Additional sensors were embedded on the backpack to retrieve environmental information including temperature/humidity (BME280 by Bosch Sensortec) and volatile organic compounds (CCS811 by AMS). Backpack activities and data during operation were recorded in the external on-board flash memory (MX25R1635F, 2 MB, Macronix) and were retrieved afterward. The insect-computer hybrid system was controlled wirelessly via Bluetooth 5.1 (2.4 GHz) using CC1352 microcontroller unit (ARM 32-bit Cortex M4F, 48 MHz, 352 kB of Flash, 8 kB of SRAM). An attached 120 mAh LiPo battery (10 × 15 mm, 2.5 g) used for power source, increasing the total weight of the whole backpack to 5.5 g but still safely within maximum payload limit of the insect [45].

Navigation Experiment

A batch of five insect-computer hybrid systems was tested in three different terrains with no obstacles, low obstacles, or tall walls (Extended Data Fig. 2). Ten trials for each system were conducted on each terrain respectively, with a five-minute interval between each trial. The experiment for each terrain was performed on separate days. The same batch of insects was used for both navigation algorithms.

Two 5-cm radius circle acted as the navigational origin and destination for the insect, which was then navigated accordingly from one circle to another. If the insect reached the destination (Extended Data Fig. 3b) within 100 s, it was counted as a successful trial. The trial was terminated for each of the following three conditions: when the insect reached the destination, when the insect was motionless for over 5 s, or when the experiment duration exceeded 100 s, whichever came first. The insect was defined to be motionless if its displacement was less than 0.5 cm [46].

The navigation experiment was conducted inside the viewable region of a three-dimensional (3D) motion capture system (Vicon®, Six T40 cameras, 4 Megapixel, 100 fps) [29,30]. A structure made of three retroreflective markers and a carbon fibres frame was attached on the wireless backpack using double sided tape to track the insect's location (Extended Data Fig. 3f) [31]. Coordinates of these markers, which represents the insect's position and orientation, were provided by the 3D motion capture system, and used to derive the motion of the insect during the navigation. Specifically, the distance ($D$) and orientation ($\Upsilon$) of the insect relative to the destination (Extended Data Fig. 3a-b) was employed in both navigation programs. They were calculated as:

$$D = \sqrt{(X_1 - X_d)^2 + (Y_1 - Y_d)^2}$$

$$\Upsilon = \cos^{-1} \frac{(X_1 - X_2)(X_d - X_2) + (Y_1 - Y_2)(Y_d - Y_2)}{\sqrt{(X_1 - X_2)^2 + (Y_1 - Y_2)^2} \sqrt{(X_d - X_2)^2 + (Y_1 - Y_2)^2}}$$

where $(X_d, Y_d)$, $(X_1, Y_1)$, $(X_2, Y_2)$, were coordinates of the destination and the two markers representing the insect's anterior and center points, respectively. The latter two information was provided by the 3D motion capture system, while the former one was predetermined by users.

A central station connected to a main PC was used to communicate with the backpack. The coordinates of the three markers were streamed to the PC, which was then sent to the central station at the speed of 30 ms for each package using a custom software written in C#. The location data was subsequently transferred wirelessly to the backpack so they could be used in the navigation feedback controls.

The two navigation algorithms (Simple Feedback and Predictive Feedback) were embedded into the backpack. These programs processed the location data and then issue the corresponding stimulation command to control the movement of the insect. The stimulation command was wirelessly transferred to the main PC via the central station for logging purposes as well as to synchronize with the location of the markers for post-experiment analysis.

Navigation Algorithms

In the Simple Feedback navigation, the orientation of the insect relative to the destination, $\Upsilon$ (Extended Data Fig. 3a), was computed every 30 ms. If $\Upsilon$ exceeded its given threshold, $\Upsilon_t$ (degree), a steering command was executed, otherwise the insect could walk freely without any stimulation.

In the Predictive Feedback navigation, a similar process occurred to decide whether the insect was (1) steered or (2) allowed to walk freely, but there were additional feedback processes for each state to predict the insect's tendency to stop and perform the required preventive measures.

(1) When the insect was steered ($\Upsilon > \Upsilon_t$) and the duration of the steering stimulation exceeded $d_s$ (ms), the insect's angular speed ($\omega$) was measured at every predetermined interval $t_v$ (ms). If $\omega$ was below its threshold $\omega_t$ (degree/s), the navigation system judged the insect to be immobile and stopped the stimulation to allow free walking for $t_{f1}$ (ms). The system then executed the acceleration command for a duration of $d_a$ (ms). The insect was then allowed to walk freely for $t_{f2}$ (ms) before the control loop returns to monitor $\Upsilon$.

(2) When the insect could walk freely ($\Upsilon \leq \Upsilon_t$), the system allowed the insect to maintain the free-walking state for $t_{f3}$ (ms) if it was steered in the loop immediately preceding the current one. The linear speed ($v_l$) was measured every $t_v$ (ms) interval. If $v_l$ dropped below a given threshold $v_t$ (cm/s), the system executed the acceleration command for a duration of $d_a$ (ms). The insect was then allowed to walk freely for $t_{f2}$ (ms) before to the control loop returns to monitor $\Upsilon$.

The control parameters including $\Upsilon_t, t_v, t_{f1}, t_{f2}, t_{f3}, d_a, d_s, \omega_t, v_t$ were unchanged throughout the experiments (Extended Data Table 1).

In the navigation experiment, $\omega$ and $v_l$ were computed based on the location data of the three markers recorded by the 3D motion capture system. To mimic real deployment conditions, the on-board IMU was used instead for the demonstration. For the former case, an average-moving filter with the window size of 250 ms was used to compute $\omega$ and $v_l$. For the latter case, $\omega$ and $v_l$ were computed using the data from the gyroscope and accelerometer. A low pass-filter of 10 Hz was implemented for noise removal.

Human Detection Experiment

A machine learning model for human detection was developed for this study. Since a low-powered processor was implemented, there were a few aspects to consider during the development. First, the memory must fit within the available space of the microcontroller, which was 1988 kB Flash and 191.8 kB Static RAM after deducting the memory required for other tasks (Extended Data Fig. 1b, Extended Data Table 3). Second, a fast-processing speed was required to achieve a real time operation. Therefore, the human detection algorithm's complexity must be low to satisfy the two requests, yet its accuracy should be high to avoid missing potential human victims.

Images used in the training of human detection model were captured in our laboratory at normal room temperature (from 25 °C to 27 °C). The thermal images of two human subjects and several non-human hot objects which varied in shapes and dimension (Extended Data Table 4) were captured using the IR camera mounted on the backpack. Each subject was placed on a turning table (BLK ND-RC6013, 120 rpm, 60 cm diameter) during the data collection process to diversify its appearance on the image. As the range of the camera was defined to be within 0.5 to 1.5 m such that the characteristics of human could be clearly seen in the captured images, the images for training dataset were captured at 0.5, 1.0, and 1.5 m. In those images containing human subjects, the number of pixels fell into the human temperature range, which was from 28 °C to 38 °C, was found as above 15 pixels. This number was then employed as a threshold to activate the on-board human detection algorithm. The collected IR images were then pre-processed and denoised using a median filter (kernel size 3 × 3). After pre-processing, the final training dataset included 11171 human images and 11493 non-human images. These images were used to train the human detection model.

The performance of the trained model was validated with a validation dataset. The dataset was prepared with a procedure mostly similar to the training dataset except for some changes made to increase the generality and variety of the images. Specifically, different human subjects and hot objects that have different shapes compared to that of the subjects used in the training data were captured (Extended Data Table 4), and the distances between the IR camera and the subjects was changed to increments of 0.1 m from 0.5 to 1.5 m.

To extract features from each thermal image, HOG was used as the feature descriptor. This method converted the distribution of local intensity gradient and the orientation into a feature vector, which was used to characterize an object's shape and appearance in the image [47]. Considering the 32 × 32 resolution of the IR images, a comparison would be made between different cell sizes used for the descriptor including 2 × 2, 4 × 4, and 8 × 8 which divided the image into 225, 49, and 9 blocks respectively, and the whole image had a feature vector length of 8100, 1764, and 324 respectively, considering that each block contained 36 features. SVM was selected as the studied classifier as its commonly used in supervised machine learning tasks [17]. The human detection classifier was then chosen among Linear SVM, Cubic SVM, and Quadratic SVM techniques. In spite of having a lower accuracy compared to its non-linear counterparts, Linear SVM has the advantage of lowest computational time owing to its simpler calculations [40,47]. Linear SVM, therefore, was employed to study the accuracy of human detection under the three different cell sizes and the cell size giving the highest accuracy would be selected for the HOG descriptor used in the final model. A comparison in accuracy and computational time between the three SVM techniques were done to select the best performing model. It is to be noted that all models used in comparisons mentioned above were trained using Matlab®. The kernel parameters of the selected model were then implemented into the backpack's main microcontroller to realize the on-board human detection system.

We used Code Composer Studio (CCS) integrated development environment to build and transfer the human detection program into the MSP432 microcontroller. To evaluate the impact of the human detection algorithm on the whole system, CCS was used to monitor the hardware resources and measure the processing time for each captured image while running human detection. The EnergyTrace tool was used to measure the power, voltage, and current during circuit operation to calculate the power consumption.

Demonstration

To demonstrate the application in search and rescue missions, an insect-computer hybrid system was navigated autonomously through an unknown environment that was made to mimic a post-disaster site (Fig. 5d, Supplementary Video 1). Several obstacles made of building materials (concrete and cement blocks) varying in shape and size were situated randomly inside a $420 \times 600$ cm$^2$ arena. From the origin, the insect was navigated through five pre-determined targets made of 8-cm-radius circles until it reached the destination, then it was directed back to the origin. The navigation was completed without sending the location information of the obstacles to the navigation system. The angular and linear speeds measurement was done using an on-board IMU to simulate the actual use case of the insect-computer hybrid system. The insect's location (X, Y, Z) was fed to the on-board navigational program via the 3D motion capture system (Vicon®). Multiple hot objects and humans were introduced nearby the navigational targets to demonstrate the performance of the on-board human detection system. The subjects were different with the one used in the data training process. IR images capture by the IR camera were stored inside an external memory embedded on the backpack. The human detection results established from these images was wirelessly streamed to and displayed on the main PC.

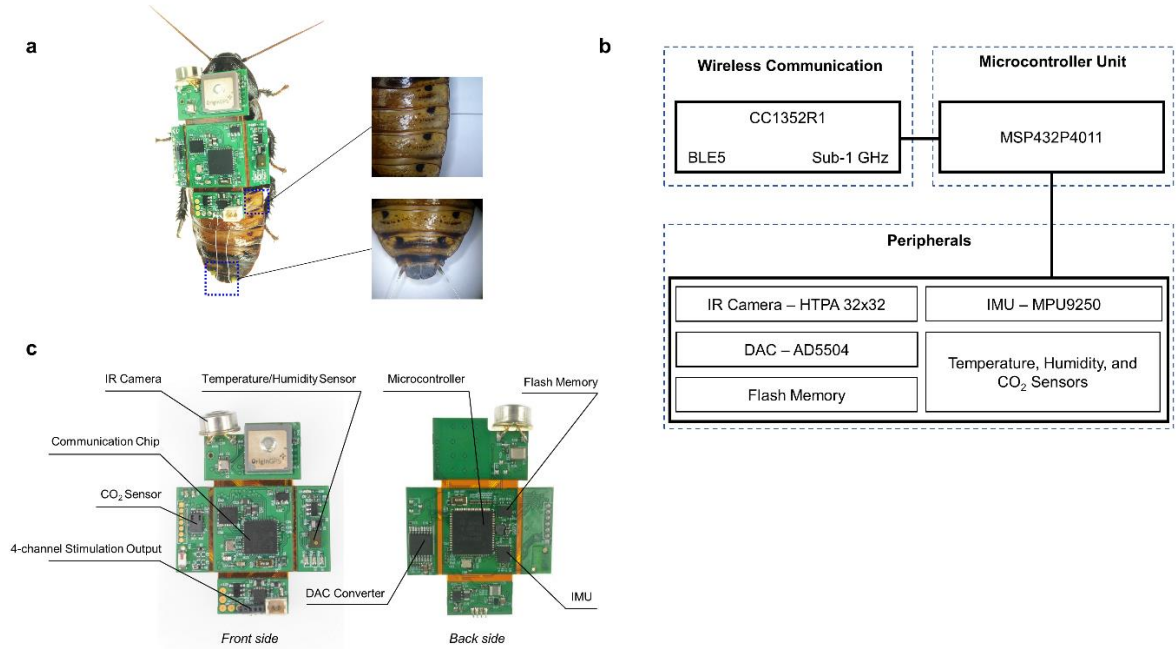

**Extended Data Fig. 1 | Overview of the insect-computer hybrid system. a,** Details of the implantation. Four electrodes in total are implanted into the insect's cerci and third abdominal segment (Right pictures). The implants are then secured with beeswax (Left picture). Electrical stimulation is released by the backpack to the insects via these wires to control its locomotion. **b,** Structure of the wireless backpack stimulator. The backpack has three functional blocks including wireless communication, main controller unit, and peripheral components. The on-board navigation and human detection algorithms are embedded into the main controller unit. Necessary data for the two algorithms is transferred to the unit via the wireless communication (location data) and from peripheral components (IR images), respectively. **c,** Overview of the backpack. To increase the backpack's flexibility, the backpack is designed as an assembly of several segments instead of a complete rigid circuit.

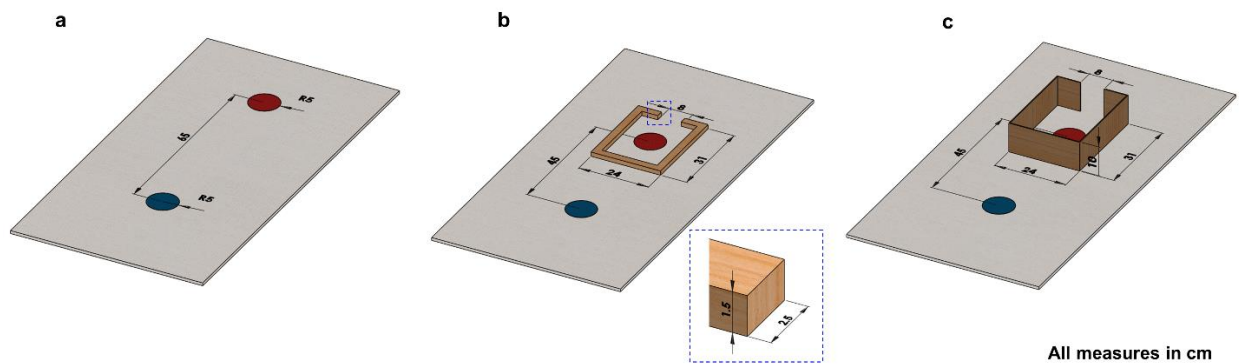

**Extended Data Fig. 2 | Terrains used in the navigation experiment.** The insect is navigated across three environments with different levels of obstruction complexity including **(a)** no obstacle, **(b)** low obstacle, and **(c)** tall wall. The terrain floor is covered with fabric and the obstacles are made of wood. The origins and destinations are each marked with blue and red circle. The distances between the two circles are set at 65 cm and 45 cm for environments with no obstacle and with obstacles, respectively. The low obstacles and tall walls are different in their height and thickness but identical in their length (31 cm), width (24 cm), and entrance's dimension (8 cm). The height of low obstacles is 1.5 cm, which is higher than the insect's usual height (~1 cm). The height of tall walls is 10 cm. The thicknesses of the low obstacles and the tall wall are 2.5 cm and 0.3 cm, respectively.

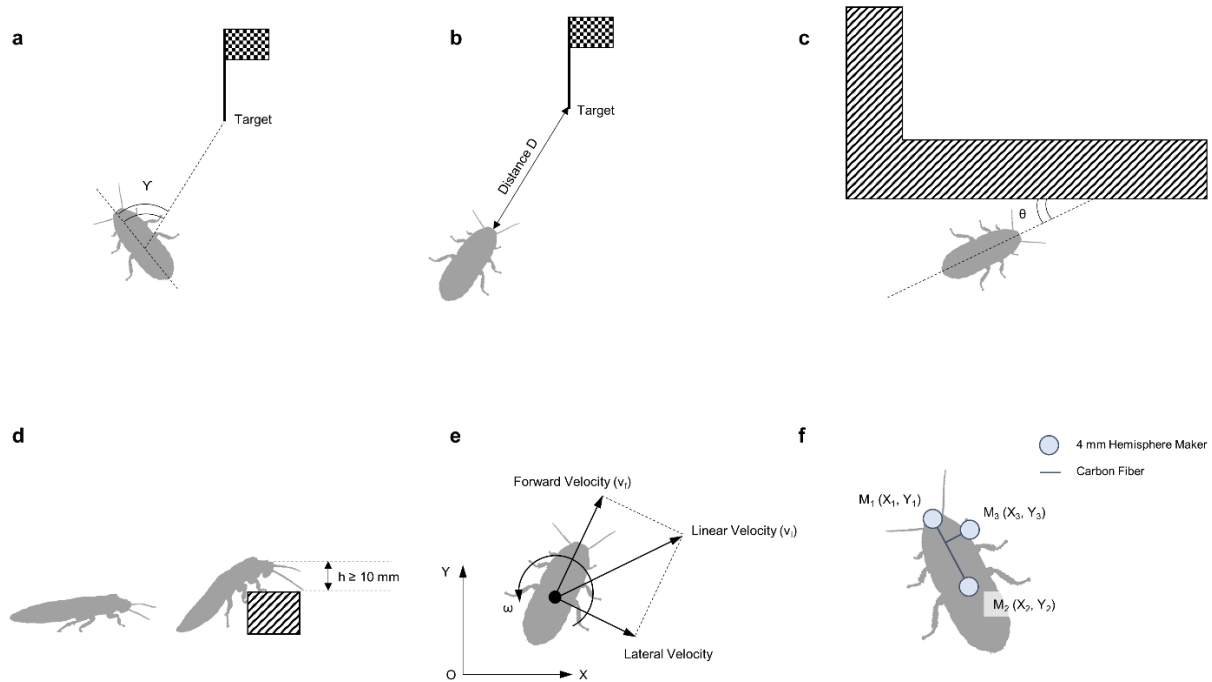

**Extended Data Fig. 3 | Graphical description of the technical terms used in the analysis. a,** The angle $Y$ represents the angular displacement between the insect's orientation and the destination, calculated based on the insect's center point. **b,** The insect's anterior point is utilized to compute the distance $D$ between itself and the destination. The insect reaches the target when $D$ is smaller than its threshold $D_t$, which is the radius of the destination region. **c,** The acute angle made by the insect's body and the nearest obstacle is named as $\theta$. **d,** The insect is judged to perform a climbing action if its anterior point is equal to or more than 10 mm above the top surface of the low obstacle. **e,** Depictions of the velocities used in the study. Angular velocity ($\omega$) is centered around the middle point of the insect. The linear velocity ($v_l$) is the superposition of two elements, which are the forward velocity ($v_f$) that is longitudinally aligned with the insect's orientation and the lateral velocity that is perpendicular with the insect's body. **f,** The anterior and center points of the insect are determined using the 3D motion capture system via three retroreflective markers. The location asymmetricity of the three markers allows the insect orientation to be recognized by the system.

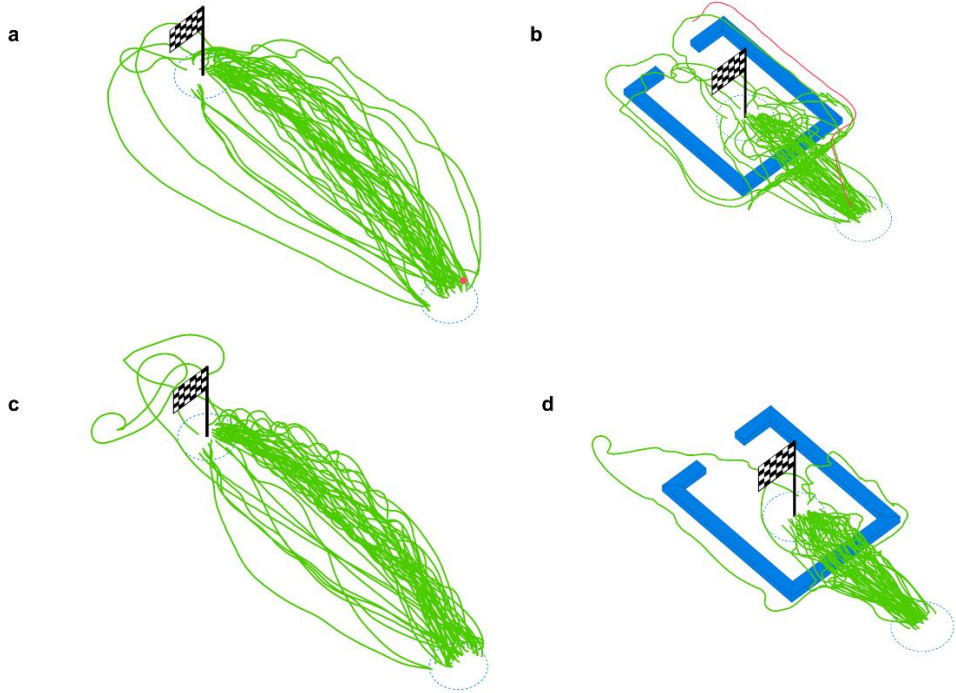

**Extended Data Fig. 4 | Trajectories of the two navigation programs for environments with and without low obstacles (N = 5 insects, 44 ≤ n ≤ 50 trials for each plot). a-b,** Trajectories obtained with the Simple Feedback navigation. **c-d,** Trajectories obtained with the Predictive Feedback navigation. Green and red curves each represent the successful and failed trials, and the red dot represents a trial where the insect is immobile. In general, the Predictive Feedback navigation retains the good results of its counterpart in these two terrains, and lowers the occurrence of immobility-related failures.

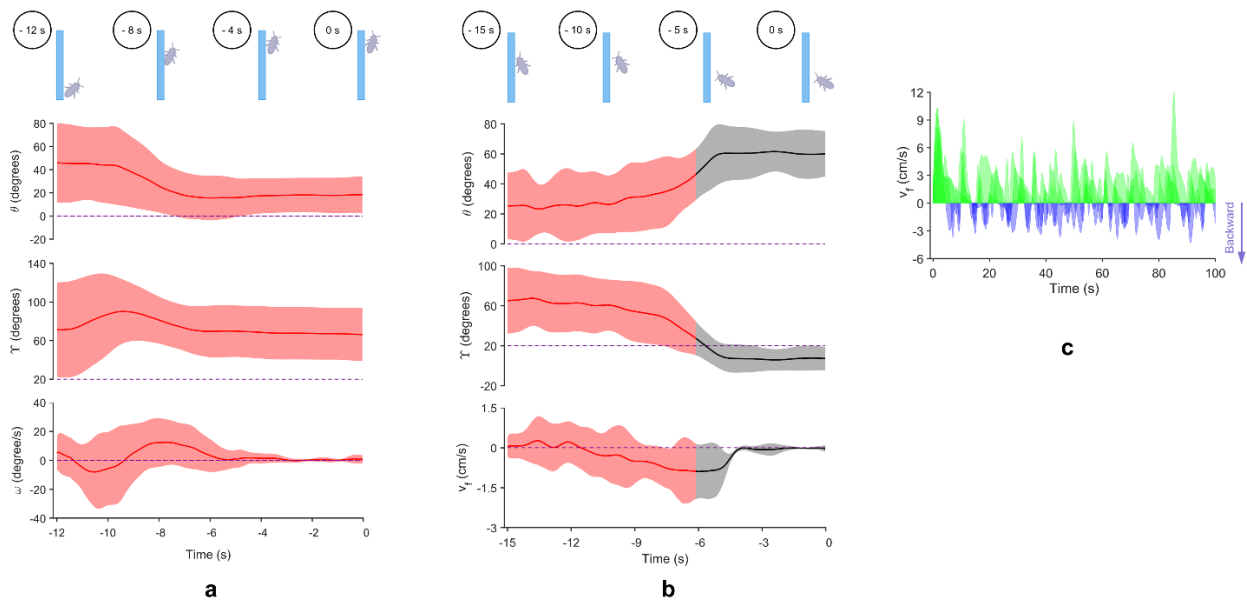

**Extended Data Fig. 5 | Detailed analysis of the three types of failure in the Simple Feedback navigation. (N = 5 insects, n = 37 trials). a-b,** The figures depicted the analysis of the immobility-related failures with the presence (17 trials) and absence (16 trials) of stimulation, respectively. The top insect-shaped objects depicted the spatial relationship between the insect and its nearest wall. The angle $\theta$, $\Upsilon$, and the speed $\omega/v_f$ of the insect are each plotted in the top, middle, and bottom charts, respectively. The horizontal axis shows the temporal information. The zeroth second is defined at the instant of experiment termination. The red and black curve each represents the mean of the variables during the stimulation and stimulation-free period. In **(b)**, the threshold time separating these two periods is the average value calculated from 16 trials. The shadow region indicates the standard deviation. The purple dashed lines represent the origin of $\theta$, $\omega$, and $v_f$ as well as the threshold of $\Upsilon$ ($\Upsilon_t$). In (**a**), the angle $\theta$ tends to reduce before remaining unchanged for the last 5 s, implying that the insect is steered to and then blocked by the obstacle. Consistently, the angle $\Upsilon$ remains nearly constant and $\omega$ tends to fade away as the insect becomes stationary. In (**b**), the insect continues to be stimulated towards the wall due to $\Upsilon > \Upsilon_t$ even if it is being obstructed (i.e., $\theta$ is small). The angle $\Upsilon$ then starts reducing until the electrical stimulation ceases, meanwhile the angle $\theta$ keeps growing bigger before remaining unchanged. These tendencies imply that, in avoiding the obstruction, the insect adjusts its orientation by performing backward motions, as seen from the negative value of $v_f$. (**c**) Analysis of the timeout failure trials (4 trials). The chart plots the $v_f$ against time in all 4 trials, with the green regions depicting the forward motion (i.e., positive values) and the blue regions depicting the backward motions (i.e., negative values). The insect spends nearly half of the navigation time moving backward to reorient its body as an act of obstacle negotiation.

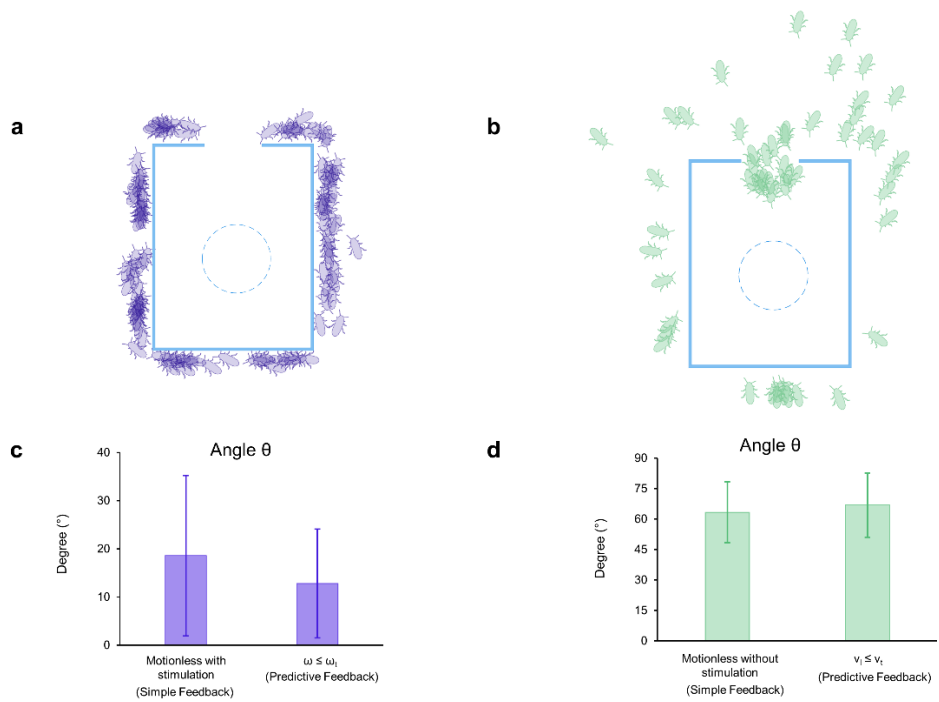

**Extended Data Fig. 6 | Effect of the Predictive Feedback navigation in failure prevention.** Graphical depiction of the insect's position when the new navigation program predicted the likelihood of failure with (**a**) small angular speed ($\omega \leq \omega_t$, purple insect-like objects) or (**b**) small linear speed ($v_l \leq v_t$, green insect-like objects). **c-d,** Comparison between the angle θ in motionless failures with and without stimulation in Simple Feedback navigation and the angle θ recorded at locations where the corresponding speed fell below threshold in Predictive Feedback navigation, specified in (**a**) – (**b**). Statistically, the similarity in both navigation program's angle θ for each failure type implies that the new navigation program successfully foresaw and prevented the potential failures that could occur with the Simple Feedback navigation.

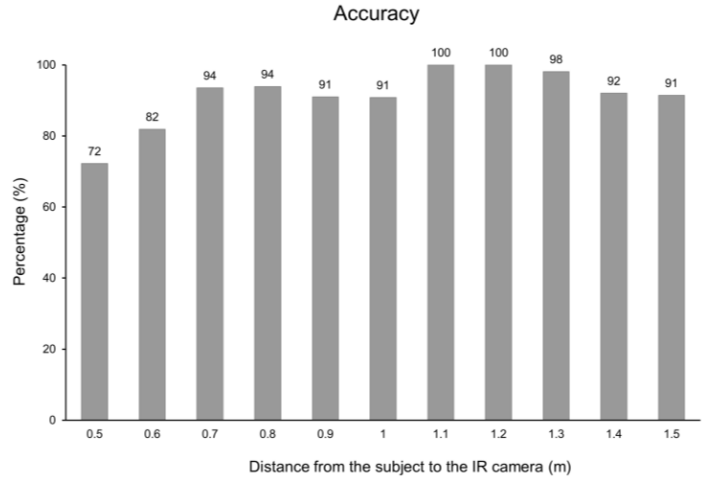

**Extended Data Fig. 7 | Validation test of the human detection algorithm.** Performance of the algorithm in accurately detecting human presence in images containing human (i.e., the rate of recall) for different distances. The dataset for this validation test was prepared with 2831 positive images (i.e., images with human subject) collected within the range from 0.5 m to 1.5 m in 0.1 m increments. The developed algorithm shows high average accuracy of 90% for subject distance within 0.5 m to 1.5 m.

**Extended Data Table 1 | Control parameters for the Predictive Feedback navigation program**

| Symbol | Definitions | Value | Unit |
|---|---|---|---|
| $Y_t$ | $Y$ threshold | 25 | degree |
| $\omega_t$ | Angular speed threshold | 5 | degree/s |
| $v_t$ | Linear speed threshold | 2 | cm/s |
| $t_v$ | Speed measurement interval | 500 | ms |
| $t_{f1}$ | Duration of free-walking state before acceleration (when $\omega < \omega_t$) | 250 | ms |
| $t_{f2}$ | Duration of free-walking state after acceleration | 500 | ms |
| $t_{f3}$ | Duration of free-walking state when $Y \leq Y_t$ and the insect was steered in the preceding loop | 250 | ms |
| $d_a$ | Duration of acceleration | 2000 | ms |
| $d_s$ | Maximum stimulation duration prior to speed measurement | 2000 | ms |

**Extended Data Table 2 | Comparison between different cell sizes used for in Histogram of Oriented Gradient method**

| Cell size | Accuracy in positive samples (%) | Accuracy in negative sample (%) |
|---|---|---|
| 8x8 | 92.2 | 80.3 |
| 4x4 | 89.9 | 84.0 |
| 2x2 | 79.0 | 79.6 |

**Extended Data Table 3 | Memory resource for human detection task and memory consumed by the developed human detection algorithm**

|  | Flash (kB) | SRAM (kB) |
| --- | --- | --- |
| Microcontroller MSP432p411 | 2048 | 256 |
| Other Tasks | 60.0 | 64.2 |
| Available Resource for Human Detection Task | 1988 | 191.8 |
| Final Developed Human Detection Algorithm | 18.3 | 52.2 |

**Extended Data Table 4 | Subjects used in the human detection development**

| Type of data | Subject | Dimension |
|---|---|---|
| Training data | Oven Cornell | 46 x 33 x 29 cm$^3$ |
| | Laptop Acer Aspire 4750G | 35 x 24 x 3.5 cm$^3$ |
| | Monitor Dell 19" | 41 x 41 x 14 cm$^3$ |
| | Human #1, #2 | Sitting height: 90 cm<br>Standing height: 166 cm |
| Testing data | Microwave Panasonic | 48 x 32 x 28 cm$^3$ |
| | Monitor HP 21.5" | 51 x 41 x 23 cm$^3$ |
| | Lamp | 44 x 44 x 16 cm$^3$ |
| | Laptop Dell Latitude 5590 | 38 x 25 x 2.5 cm$^3$ |
| | Human #3 | Sitting height: 90 cm<br>Standing height: 166 – 172 cm |

Supplementary Information for

# Insect-Computer Hybrid System for Autonomous search and rescue mission

P. Thanh Tran-Ngoc, D. Long Le, Bing Sheng Chong, H. Duoc Nguyen, V. Than Dung, Feng Cao, Yao Li, Kazuki Kai, Jia Hui Gan, T. Thang Vo-Doan, T. Luan Nguyen, and Hirotaka Sato*

*Corresponding author. Email: hirosato@ntu.edu.sg

**Autonomous navigation of insect-computer hybrid system and human detection in an unknown mock-disaster environment**

**Video 1.** The insect's motion is automatically directed by the on-board navigation program. The program maneuvers the insect from the origin to the predetermined destinations without having information about the location of surrounding obstacles. The navigation can be successfully achieved with the cooperation between the navigation program and the insect's natural locomotive ability. To aid in survivor searching, the insect-computer hybrid system is equipped with an IR camera that captures images at 1 Hz and is suitable for operation in dark areas. By processing the collected IR images, the on-board machine learning (ML) algorithm is able to correctly identify human presence among different human/non-human hot subjects. The results are wirelessly reported to the command center in real time.

**Locomotion control of insect-computer hybrid system**

**Video 2.** The insect turns left, turns right (first half of the movie) or accelerates (second half of the movie) under the user's commands. The commands are wirelessly transferred from a computer to the backpack stimulator mounted on the insect's body. The backpack then interprets the received commands to generate the corresponding electrical stimuli, which are applied to the insect to induce the desired motions.